# Reinforcement Learning for Robotics and Control with Active Uncertainty Reduction

Narendra Patwardhan[1], Zequn Wang[2]

**Abstract** — Model-free reinforcement learning based methods such as Proximal Policy Optimization, or Q-learning typically require thousands of interactions with the environment to approximate the optimum controller which may not always be feasible in robotics due to safety and time consumption. Model-based methods such as PILCO or BlackDrops, while data-efficient, provide solutions with limited robustness and complexity. To address this tradeoff, we introduce active uncertainty reduction-based virtual environments, which are formed through limited trials conducted in the original environment. We provide an efficient method for uncertainty management, which is used as a metric for self-improvement by identification of the points with maximum expected improvement through adaptive sampling. Capturing the uncertainty also allows for better modeling the reward responses of the original system. Our approach enables the use of complex policy structures and reward functions through a unique combination of model-based and model-free methods, while still retaining the data efficiency. We demonstrate the validity of our method on several classic reinforcement learning problems in OpenAI gym. We prove that our approach offers a better modeling capacity for complex system dynamics as compared to established methods.

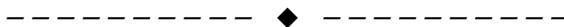

## 1 INTRODUCTION

CURRENT reinforcement learning algorithms often require a large number of trials to autonomously learn a controller, due to the high dimensional continuous nature of the state space, thereby prohibiting them to be a standard tool in robotics. Conducting trials in the real world to obtain sample data is exigent due to real-time nature and potential for mishaps due to incorrect policies, making data efficiency a critical decisive factor while choosing a method for controller formation. Simulation models allow for exploration of dynamic nature of the world without these limitations, augmenting and providing prior knowledge about the environment without the involvement of any hardware, however they may be costly to emulate depending upon the complexity and fidelity of simulation. For robotics and control, the performance and the robustness of the controller is highly linked with the fidelity of simulator, leading to requiring higher computational resources than many other domains.

Model-free methods such as Q-learning rely on experience replay and therefore require a large number of trials. Creation of a surrogate model to approximate the true state transition function for the Markov Decision Process (MDP) problem or the model-based approach allows for prior predictive planning thereby reducing the number of trials significantly. However, they introduce model uncertainty due to lack of data which propagates through the state transitions and remains intractable in the output. Probabilistic models of state transition not only provide for the prediction of next state but also the uncertainty about the prediction based on known data which can be taken into account while designing the control policy. The primary focus of this work is to provide a systematic method to quantify and reduce uncertainty for robust controller design, which we call 'Active Uncertainty Reduction' (AUR).

PILCO [1] or probabilistic inference for learning control provides one such method that provides a posterior state distribution as the output of state transition function which acts as prior for next prediction. This is achieved through the simplifying assumption of moment matching, which approximates the next state as a normal distribution. However, this assumption may not always hold true depending upon the complexity of the system dynamics. Modern computational systems allow for efficient tensor calculations through function vectorization, which makes a sampling-based realization of state transition a much more desirable alternative as compared to losing information through distribution approximation. Since sampling based posterior distribution consists of individual travels through a finite number of state distributions, it allows for computation of gradient information through an Autograd framework which is the basis of most modern neural network frameworks [2] instead of requiring the computation of analytical gradients, opening avenues for more versatile policy structures.

While PILCO embeds the model uncertainty in the planning or optimization processes, simulation environments provide an opportunity for uncertainty reduction prior to the optimization step through explicit identification of the points in state space which would improve the performance of the model. This method of computing expected improvement goes back to [3] and was brought to mainstream attention by [4]. Thompson sampling has been used in recent RL literature to reduce the number of observations while improving the performance of the controllers for model-free methods. Adaptive sampling, in a similar fashion, provides for a way to drastically reduce the posterior uncertainty for model-based methods. The quantification of model fidelity in terms of uncertainty information is a challenging task required for adaptive sampling. Through this work, we provide a framework which can effectively quantify and reduce the uncertainty for model-based methods.

---
[1]*Graduate student, Michigan Technological University, Houghton, MI 49931. Email: narendra@mtu.edu*
[2]*Correspnding author, Michicgan Technological University, Houghton, MI 49931. E-mail: zequnw@mtu.edu*

In this paper, we introduce a method to leverage simulation and uncertainty quantification without making any assumption about the nature of the system dynamics to learn a controller in a data efficient way. Our approach starts with an online phase to evaluate an extremely low number of simulations for data generation. With the training data, both the transition dynamics as well as the reward function are approximated by utilizing Gaussian Processes (GPs) which are a form of nonparametric regression. After the creation of these low fidelity models, they are augmented in a principled way through adaptive sampling. After improving the fidelity of our models, in the offline phase, we can optimize the controller policy using any of the model-free methods established in the literature. In this work, we pair our method with proximal policy optimization due to its applicability to both continuous and discrete action problems. After the training, we evaluate our models in the original environment.

This article first presents a survey of related work in Sec. 2 and then the framework for modeling using Gaussian Processes and calculation of expected improvement for adaptive sampling procedure is explained in Sec. 3. Sec. 4 delves deeper into the uncertainty-based reward adjustment to minimize effect of modelling error and policy optimization. We present the experimental results on two different classical reinforcement learning problems in Sec. 5 with Sec. 6 concluding the paper with key properties of the active uncertainty reduction (AUR) algorithm and a brief discussion on scaling to higher dimensional problems.

## 2 RELATED WORK

Model-based methods in reinforcement learning follow two primary ways to model the state transition function, namely, LSTM (Long Short-Term Memory) which is a form of recurrent neural network and Gaussian processes. [5], [6] use LSTMs to provide the approximation of the next state, while [7], [8], [9] use it to calculate the Q function or value function of the next state. Performance of LSTMs like all neural networks depends on the amount of data to a large extent, requiring an extensive amount of training time as well as simulation runs. LSTMs need to repeat this costly training procedure every time a significant amount of new data is obtained through the experiments. In addition, LSTMs provide a deterministic prediction and no information about the uncertainty of the prediction. In the absence of sufficient information about the state-space, this deterministic nature of prediction, adds more uncertainty at each predictive state, and accumulates errors for long-term predictions, making LSTMs unsuitable for long-horizon predictions.

Gaussian processes have been used extensively in the domain of reinforcement learning as well as robotics for modeling the state-space. PILCO [1] and it's derivative methods such as [10], [11] utilize Gaussian processes with moment matching to predict next state distribution. In classical robotics, [12] uses GPs to model continuous time trajectories with start and end states fixed for efficient motion planning. [13] utilizes GPs to provide occupancy maps for range sensing robots. Gaussian processes excel at prediction with a low amount of data however the training time increases significantly with the number of data points due to the $O(n^3)$ cost aassociated with inversion of matrix required in the fitting process. Sparse GP models were developed in [14] and [15] for this purpose which work well on larger datasets by considering a fraction of dataset as an anchor. The same problem is also tackled by [16] with intentions of developing global inverse dynamics model for the system by creating multiple local Gaussian process regression models only the closest of which is used for prediction, thereby reducing the number of points each model has to train with. One of the drawbacks that remains with the use of Gaussian processes is the inability to scale up with dimensionality. However, this problem is associated with most modelling techniques and is usually solved using dimensionality reduction techniques such as principle component analysis.

In recent literature, Bayesian Neural Networks or BNNs were proposed as an alternative to Gaussian process-based modeling in PILCO in [10]. This approach, while requiring a slightly higher amount of data than [1], retains the non-linear nature of GPs as well as provides probabilistic predictions with uncertainty information. However, BNNs are sensitive to the weight initialization methods and could be tricky to converge.

The closest approaches to our method are BlackDrops [17] and World Models [18]. [17] also utilizes a Monte Carlo sampling-based technique for finding the posterior distribution. However, no effort is made prior to optimization to reduce the variance of posterior instead of relying on data obtained by performing multiple random rollouts. Our method relies on the technique of adaptive sampling which has shown prominent use in the field of reliability based design and optimization [19], [20] and systematic variance reduction. [18] uses a surrogate environment also termed as dream environment to train the policy in and evaluates its performance in the real environment, however, the surrogate model is created via the use of LSTMs with $10^5$ trials, thus being orders of magnitude less data-efficient than other model-based methods. We treat each simulation step as an uncertain estimate of the true state, which is a more appropriate estimation of system uncertainty. The policy optimization in [17], [18] relies on the population-based approach CMA-ES, which can be significantly slower than gradient-based approach without utilizing parallel processing if the gradient-based method can find a solution. Our approach utilizes the same optimization routine described in [21]. Since our method can easily be combined with gradient-based optimization methods, it is more suitable for use with neural network controllers with higher policy parameters than Blackdrops with CMA-ES, as evolutionary algorithms have been shown to be significantly harder to converge for neural networks for more than a couple of layers due to exponential increase in the dimensionality. While [1] and its derivatives rely on improving model fidelity by incorporating more data through conducting trials, an intrinsic way to determine which data to include for GPs is adaptive sampling which has been used for design optimization by [22], [23] and in the field of black box function optimization by [24]. Adaptive sampling to increase the fidelity of the Gaussian process has been used in [25], [26]. While several variants of adaptive sampling rely on various sources of uncertainties, we combine modeling uncertainty with objective function metric to have efficient exploration and identify not only the areas in which state transition model needs improvement but also where it is most useful to update to help policy converge.

## 3 VIRTUAL ENVIRONMENT MODELLING WITH ACTIVE UNCERTAINTY REDUCTION

In the following section, we describe the system under consideration and the major components of the virtual environment creation procedure as shown in Fig.1. We introduce Gaussian process regression and show how in combination with Latin hypercube sampling, they can be used for creation of low fidelity virtual environment. We then delve into how this fidelity can be improved sequentially in a Bayesian manner with the help of active uncertainty reduction procedure.

The behavior of an environment can be completely described by two functions, state transition function and reward function. Together these function form a step function which takes an action as input and changes the state of an environment to another state and outputs a reward for being in the new state. Optionally, it provides additional information on whether the objective has been met or not. To create the equivalent virtual environment, we need to define these two functions. Another optional parameter in defining the environment is the starting state distribution which defines the placement of the agent(s) at the start of each run of the experiment. The starting state distribution varies from environment to environment and in this work, it is defined using the prior knowledge of the system, however, if none is available, a uniform or uninformative prior can be assumed for each state dimension.

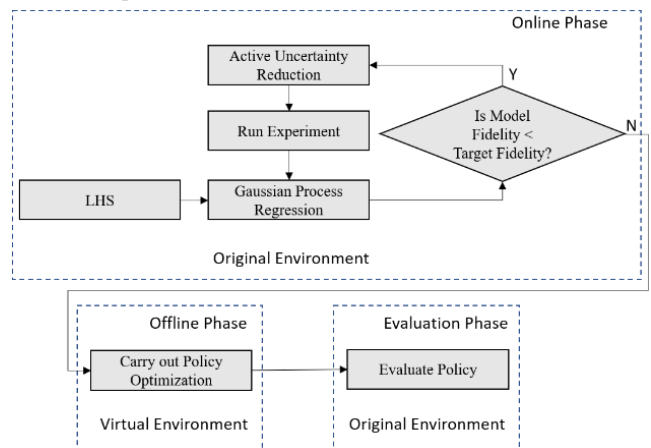

Fig. 1. Overview of the AUR algorithm.

Carrying out each step of simulation is costly as it usually involves solving differential equation(s). Real world experiments are similarly costly in terms of energy and risks associated. Creation of a virtual environment, which can mimic the response of the original system, at a lesser computational cost is thus often desirable. The fidelity of the virtual environment is highly dependent of data used to create it. We collect the initial data using Latin Hypercube Sampling (LHS) as opposed to random rollout as in [1], [17] & [18] to have a balanced representation of the state-space. A low fidelity virtual environment is created using GP modelling. The fidelity of model is increased using active uncertainty reduction which we outline in section 3.3. After the target fidelity is achieved policy optimization takes place in the virtual environment.

Depending upon the target fidelity and the computing power available, we provide two variants of transition in the virtual environment. A deterministic variant (AUR-D) in which only a single internal state is stored at any time instant and only the prediction mean through state transition approximation is used for calculation of reward. In the probabilistic or sampling-based approach (AUR-P), multiple states ($10^3+$) are stored internally at each instant, while at each instant a representative state is provided using the mean of all stored states, the reward approximation depends upon all states. The transition flow in both AUR-D and AUR-P is outlined in Fig. 2.

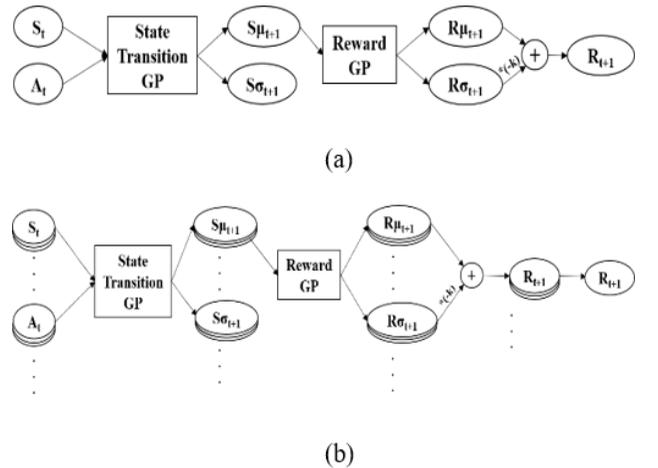

Fig. 2. Reward prediction in virtual environment (a) Deterministic variant, (b) Probabilistic variant.

### 3.1 Gaussian Process Approximation of Transition and Reward Function

We consider our environment to be based on a dynamic system with continuous states $x \in \mathbb{R}^D$ and controls $u \in \mathbb{R}^F$, with unknown transition function $f$ described as:

$$x_{t+1} = x_t + f(x_t, u_t) + w \tag{1}$$

$$w \sim \mathcal{N}(0, \Sigma_w) \tag{2}$$

where $w$ is i.i.d. Gaussian noise. Our objective is to find a policy $\pi: x \mapsto \pi(x, \theta) = u$ such that it would maximize the expected long-term reward over a fixed horizon $T$.

$$J^\pi(\theta) = \mathbb{E}[\sum_{t=0}^{T} r(x_t, u_{t-1}) | \theta] \tag{3}$$

where we define $r$ to be some reward function dependent upon the current state $x_t$ and the action taken to reach the state $u_{t-1}$. We approximate the state transition function $f$ using independent Gaussian process regression model for each target dimension $[f_s^1 .. f_s^D]$. Gaussian processes is a type of nonparametric regression model, which can forecast unknown response of new data after approximating the model parameters through the use of initial training samples. A Gaussian process is completely describable by its mean function and covariance function. For states $x = [x^1, x^2, .., x^D]$ and control $u = [u^1, u^2, .., u^F]$, input for the GP is $\tilde{x} = [x, u]$, we approximate each state for the next time instant as:

$$x^i = f_s^i(\tilde{x}) = \mathcal{GP}_i(\tilde{x}) = \mu_i(\tilde{x}) + S_i(\tilde{x}) \qquad (4)$$

where for the sake of simplicity $\mu_i$ is taken as a constant function with zero as output and the covariance function of $S(\tilde{x})$ given by

$$k\big(S(\tilde{x}_p), S(\tilde{x}'_q)\big) = \sigma_f^2 e^{-\frac{1}{2}(\tilde{x}-\tilde{x}')\Lambda(\tilde{x}-\tilde{x}')} + \delta_{pq}\sigma_n^2 \qquad (5)$$

The covariance function or kernel chosen is the squared exponential kernel with automatic relevance determination. The GP hyperparameters are the diagonal entries of the matrix $\Lambda$ (length scales) and signal variance $\sigma_f^2$ and the noise variance $\sigma_n^2$, $\delta pq$ is 1 if $p = q$ and 0 otherwise. With the hyper parameters trained by maximizing the likelihood function, the mean and variance of the GP prediction are given as

$$E_f(\tilde{x}_t) = k_* \beta \qquad (6)$$

$$Var_f(\tilde{x}_t) = k_{**} - k_*(K + \sigma_w^2 I)K_* \qquad (7)$$

where $k_* = k(\tilde{X}, \tilde{x}_t), k_{**} = k(\tilde{x}_t, \tilde{x}_t)$, and $\beta = (K + \sigma_w^2 I)$ in which $K$ is the kernel matrix with entries $K_{ij} = k(\tilde{x}_i, \tilde{x}_j)$. We follow the heuristic guideline for the number of initial training points as 10 times the input dimensionality for the GP [19]. The reward function is similarly modelled using a GP with zero mean function and squared exponential kernel with automatic relevance determination as covariance function. This procedure when combined with the Latin hypercube sampling described in the next section allows for creation of a low fidelity virtual environment.

### 3.2 Virtual Environment Initialization with Latin Hypercube Sampling

Latin hypercube sampling is an efficient sampling method which divides a range of each dimension of the environment in the desired number of equally probable intervals. From each axis aligned hyperplane, only one sample is taken. This allows for more efficient exploration of the state-space as opposed to taking completely random samples. [1], [17] rely on generating a rollout with starting point drawn from a normal distribution with mean as default reset position and fixed standard deviation proportional to the scale of each dimension. This rollout highly depends on the starting random policy and may lead to training a GP whose prediction accuracy is highly localized. During to policy update this could lead to mispredictions.

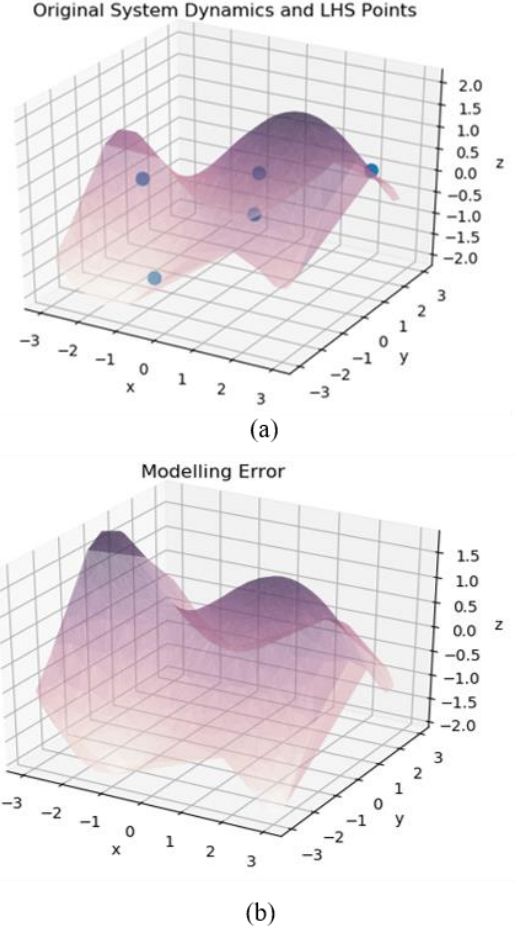

Fig. 3. (a) Original system dynamics & (b) Error for low fidelity model.

To carry out LHS for creation of low fidelity GP models we rely on previous knowledge of the state action space, i.e., we assume the bounds and distribution of each dimension for the states $x \in \mathbb{R}^D$ and controls $u \in \mathbb{R}^F$ are known. If the distribution is unbounded, we use a truncated version of the distribution. LHS is then carried out on the combined state-action space $\tilde{x} = [x, u]$ by dividing it in N equally probably hypercubes and taking a single random sample from each hypercube. Hence entire state-action space is covered. Thus, LHS allows for the creation of a low fidelity global model that is equally performant for the entire state space and whose local performance can be improved by sequentially adding more points from the region under consideration as per the acquisition function as shown in Fig 3.

### 3.3 Uncertainty Reduction using Adaptive Sampling

The global low fidelity surrogate model obtained from GP(s) may lead to high variance depending upon the states involved at each step prediction. To reduce this uncertainty more training data is needed.

After fitting the initial GP models, $N$ Monte-Carlo samples are generated denoted by $\tilde{X}$. A cumulative measure of model certainty termed here onward as Confidence Index (CI) can be calculated by the arithmetic mean of model certainty for all the samples in $\tilde{X}$.

$$CI(\tilde{X}) = \sum_{i=1}^{N} c\_m(\tilde{x}\_i)/N \tag{8}$$

$$C_m\left(\mathcal{GP}(\tilde{x}_{*,t})\right) = \phi\left[\frac{1}{\sigma(\mathcal{GP}(\tilde{x}_{*,t}))}\right] \tag{9}$$

This value (CI) is always bounded between [0.5, 1] and hence can be used as a measure of fidelity of virtual environment to the original environment. For a high-fidelity virtual environment, this value is recommended to be greater than or equal to 0.99. Higher values of CI indicate better predictive ability at a global level.

We now identify the sources of uncertainty involved which are dominant in the performance of GP models. The inherent model uncertainty involved due to the use of Gaussian processes is given as

$$U_m\left(\mathcal{GP}(\tilde{x}_{*,t})\right) = 1 - \phi\left[\frac{1}{\sigma(\mathcal{GP}(\tilde{x}_{*,t}))}\right] \tag{10}$$

The input uncertainty or the probability of a state materializaing during the run of experiment ($P_i$) can be quantized by fitting a probability distribution over the samples and is given by

$$P_i\left(\mathcal{GP}(\tilde{x}_{*,t})\right) = p(\sum_{t=0}^{T} \mathcal{GP}(\tilde{x}_{*,t})) \tag{11}$$

The outputs of GP may or may not follow a specific distribution depending upon the original system and hence using the normal distribution as in [1] to approximate would lead to errors. We hence utilize kernel density estimator as a technique to find the probability.

We introduce the uncertainty index $U(\tilde{x}_{*,t})$ to identify the failings of the global model as the function of $U_m$ and $P_i$ as

$$U(\tilde{x}_{*,t}) = U_m(\tilde{x}_{*,t}) * P_i(\tilde{x}_{*,t}) \tag{12}$$

We identify the sample with maximum expected improvement as the one with maximum uncertainty involved and call it an adaptive sample.

$$\tilde{x}_{adaptive} = \text{argmax}_t(U(\tilde{x}_{*,t})) \tag{13}$$

After the calculation of the CI values for all the GPs, a single adaptive sample is found for each GP if CI value is lower than the target value. The response for this sample is evaluated using the original environment and all the GPs are updated with this response. Thus, the uncertainty index acts like an acquisition function for Bayesian updating as described in [27]. The adaptive sampling procedure is outlined in Fig. 4. This procedure is repeated till state and reward GPs with CI greater than target are obtained, after which they can be used to emulate the step function. The prediction of reward GP is shifted to its lower confidence bound by subtracting $k$ times the standard deviation of the prediction from it as shown in Eq. 14. The k value in our experiments was kept as 3.

$$R = R_\mu - kR_\sigma \tag{14}$$

This method gives a pessimistic approximation of the reward. If the uncertainty at the prediction is extremely low, then this corresponds to using the prediction as it is, however, if the prediction uncertainty is high, this leads to lower reward than in the original environment. Like [19], optimization carried out on this significantly difficult environment leads to optimized results in the original environment.

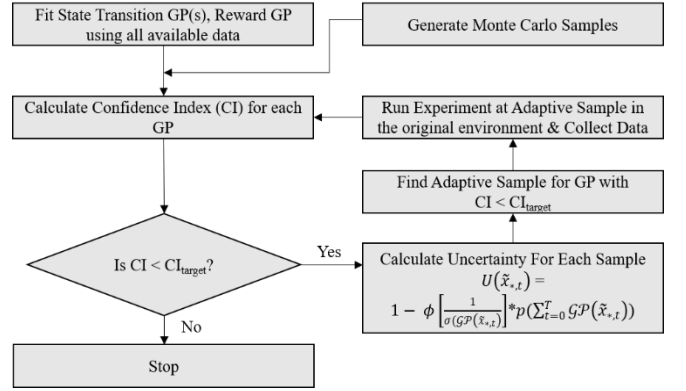

Fig. 4. Adaptive sampling procedure.

The step function for the virtual environment is defined using the state transition GP and reward GP once the CI value reaches the target CI value. The overall algorithm is described in Alg 1.

**Algorithm 1** VIRTUAL ENVIRONMENT CREATION

1: *Generate* $10(D + F)$ initial data points $\tilde{X}_{lhs} = [X\ U]$ using LHS where $X \in \mathbb{R}^D, U \in \mathbb{R}^F$
2: *Run* a single step on $\tilde{X}_{lhs}$ in original environment
3: *Fit* State Transition GP $f_s$ using all data
4: *Fit* Reward GP $f_r$ using all data
5: *Generate* Monte Carlo data $\tilde{X}_{mcs}$
6: *Calculate* CI for each GP with $\tilde{X}_{mcs}$
7: **repeat**
8:     **for each** GP with $CI < CI_{target}$
9:         *Calculate* Uncertainty Index $U$ for $\tilde{X}_{mcs}$
10:         *Select* adaptive sample $\tilde{x} = \text{argmax}(U(\tilde{X}_{mcs}))$
11:         *Run* a single step on $\tilde{x}$ in original environment
12:         *Update* All GPs using all data
13:     Calculate CI for each GP with $\tilde{X}_{mcs}$
14: **until** $CI \geq CI_{target}$
15: *Define* step function for Virtual Environment using GPs

## 4 POLICY STRUCTURE AND OPTIMIZATION

Due to the equivalent behavior as the original environment, both gradient based and gradient free optimizers can be used for policy optimization. Autograd framework [2] can be used to compute the gradients w.r.t. the policy parameters directly. Hence our approach allows for utilizing neural networks as policy structures. In our experiments, we utilize a simple form of feedforward neural networks with 1 hidden layer, composed of 32 neurons. For all the tasks we utilize the number of state dimensions as the number of input neurons. We provide two different structures for the output layer, depending upon the nature of control.

For continuous control tasks, the number of output neurons is kept equal to the number of output dimensions. We utilize hyperbolic tangent as the activation function after each layer, this maintains the output at each stage in the range (-1, 1). We then multiply the output of the final layer with a vector $u_{\text{maximum}}$ to scale it to the action space.

ReLU or Rectified Linear Unit is used as the activation function for the hidden layer in case of discrete control. There are a number of possible actions for each output dimension. We thus have $\sum_{i=0}^{E} a_i$ neurons in the output layer, where $a_i$ is the number of discrete actions possible for output dimension $i$. The action probabilities are computed using dimension-wise Softmax function and output is assigned with action having maximum probability in each dimension. To utilize a uniform controller structure, a continuous control task can also be converted to a discrete control task by discretizing the action space into finite a number of actions. This through our experiments has shown to decrease the learning time needed and can be an effective way for reducing the space complexity that problems in robotics usually suffer from.

In the deterministic variant of the algorithm AUR-D, only a single state $x_t$ is stored at a time. The action $u_t$ to be taken at this state is calculated using the policy. The next state $x_{t+1}$ is calculated using this state and action with state transition GP. This state is utilized to calculate the reward $r_{t+1}$ using the reward function GP. The reward is then penalized using the prediction variance as described before in Eq. 14. This simplistic nature of reward prediction leads to faster computation time, however if, the uncertainty involved in state prediction is high, this could lead to large bias in the reward value.

The probabilistic variant AUR-P stores several internal states ($10^3+$) and a representative output state. At the start of each run, an initial state distribution is obtained from the reset state using a fixed low variance for each dimension as described in [1] and the reset state is kept as the output state. The state prediction at each subsequent state is obtained by predicting the next state for all internal states and output state is chosen as the arithmetic mean of the predictions. Similarly, the reward is calculated for all internal states, adjusted using the Eq. 14. This leads to a reward distribution. The reward for the state is found out as the mean of the predictions weighted with $1/U$, where U is the total uncertainty involved in prediction given by equation 7. This leads to a better approximation of the reward at each instant.

We use the PPO (Proximal Policy Optimization) method which is the first order stochastic gradient-based method suitable for both continuous and discrete-valued problems to optimize the policy parameters. One of the advantages of PPO is the ability to scale well with the number of policy parameters. Unlike [17], which requires multiple CPU cores to speed up the computation, our method can take advantage of the GPU architecture which has been an established method of speeding up neural network related.

## 5 EXPERIMENTAL RESULTS

In this section, two case studies of classical control problems, Inverted Pendulum and Cart pole Balancing evaluated on the modern context [28], are used to demonstrate the efficiency of our approach. The use of OpenAI Gym environments instead of having fixed plant models allows for more complex cost or reward functions to be created, which are not only based on the states but also on the actions taken as well as the cumulative propagation effect. This permits for more complex behaviors to be achieved through the optimization process and for ease of comparison against other approaches. OpenAI gym uses reward functions, which can depend upon a variety of factors, including but not limited to last state, last action, current state and time steps taken since the start but it is quite straightforward to find a simplified version using only the current state and action taken to achieve the current state.

To test the performance of the policy purely learned through the simulation we change the system parameters by adding Gaussian random noise, which introduces the bias between the simulation and the real environment. Due to adjusting the reward response in a pessimistic way, it is significantly harder to perform well in the virtual environment as opposed to original environment, and hence the policies learned show significant robustness without needing to be retrained or calibrated. The target CI for all the experiments was kept as 0.999. The total number of interactions in the virtual environment were capped at $10^6$. All the experiments were run on system with core i7-8 generation CPU with 64 GB ram and NVIDIA Quadro P2000 GPU.

### 5.1 Inverted Pendulum

The inverted pendulum problem consists of a freely swinging pendulum with length 1 m and mass 1 kg. The objective is to learn a controller which can keep the pendulum vertical (at 0°) with least angular velocity and least effort. Unlike classical implementations of this problem where the goal is to minimize the weighted quadratic difference between state and the target state, OpenAI gym considers the action taken to reach the state in consideration as well, making it a more practical orientated composition of the problem. Due to the continuous nature of the control, the problem remains challenging especially for methods involving gradient-based optimization.

The inverted pendulum problem has a continuous observation space of dimensionality 3. The first two dimensions correspond to the physical coordinates of the end point of the pendulum $(l\cos(\theta), l\sin(\theta))$ where $\theta$ is normalized between $[-\pi, \pi]$ internally. The third state is the angular velocity $\dot{\theta}$ which is limited between [-8, 8]. The control action $a$ is the continuous variable of the joint effort, which ranges between the limit [-2, 2].

Since this simulation environment is computationally quite cheap to solve, original simulation is carried out and the response error is compared with virtual environment at each stage of adaptive sampling in Fig. 5, which shows the confidence index shows increase while error is decreased with inclusion of more adaptive samples. The low fidelity environment was create using 40 initial points with reward CI as 0.64 and total of 125 evaluations in the original environment were needed to construct the high-fidelity virtual environment as shown in Fig. 6. Fig. 7 shows the error in reward response of the low fidelity virtual environment, which differs grossly from the original environment. It should be observed however that due to the reward shaping as given by Eq 13, the reward response is always lower than the reward response of the original environment. Fig. 8 shows after the target fidelity level has been achieved, the original system response is tracked closely by the virtual environment's response.

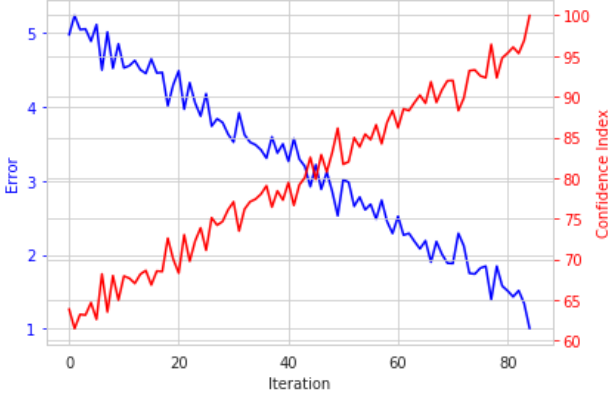

Fig. 5. Effect of adaptive sampling on Confidence Index and error.

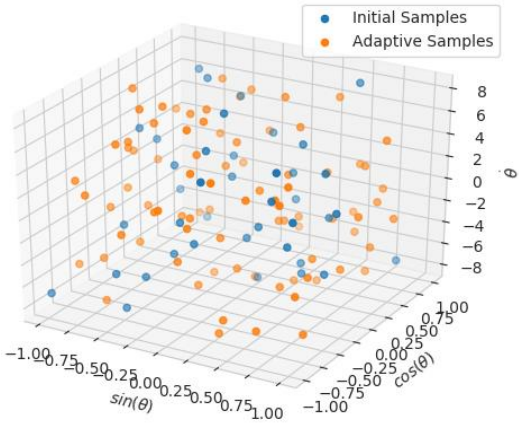

Fig. 6. Initial and adaptive samples for creation of virtual environment

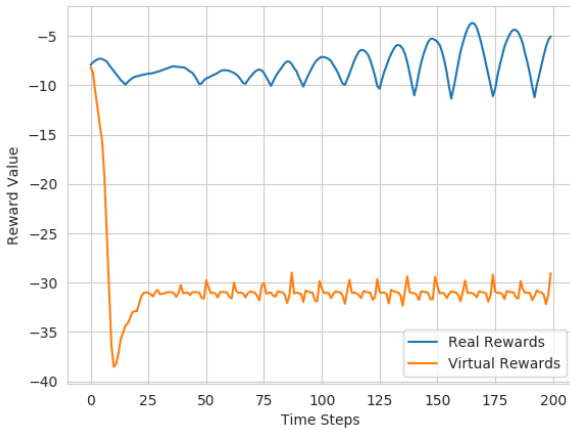

Fig. 7. Difference in reward response for low fidelity virtual environment.

The loss at a particular state to keep the pendulum in any target state with a set effort is calculated as follows to mimic the reward function used internally.

$$L(\tilde{x}) = (\theta - \theta_t)^2 + 0.1(\dot{\theta} - \dot{\theta}_t)^2 + 0.001(a - a_t)^2 \quad (15)$$

Here the targets $(\theta_t, \dot{\theta}_t, a_t)$ were (0, 0, 0). As is evident from the limits of the state-space, the theoretical minimum possible reward is -16.2736044, whereas the theoretical maximum possible reward is 0. There is no specific termination criteria specified for the task, and specifying maximum steps is suggested. Each step taken in the environment corresponds to 0.05 seconds. The maximum number of steps for simulation are set at 200, which corresponds to 10 seconds.

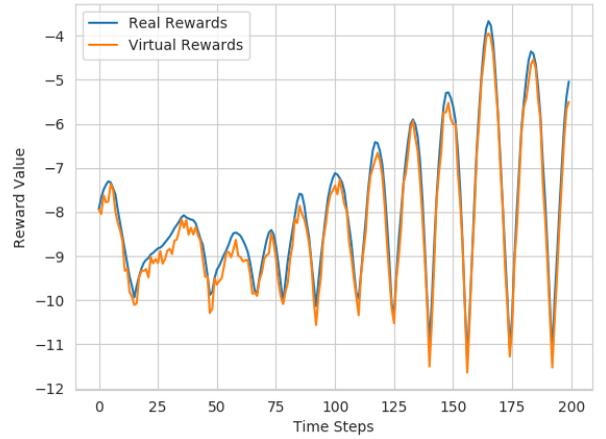

Fig. 8. Difference in reward response for high fidelity virtual environment.

The controller successfully achieves convergence in less than 4 seconds on an average. As this environment has a well-defined target state, the results are described in terms of a loss value rather than the reward value. Fig 9 shows the change in loss value across the run of the experiment on 5 random initial states by applying the learned controller.

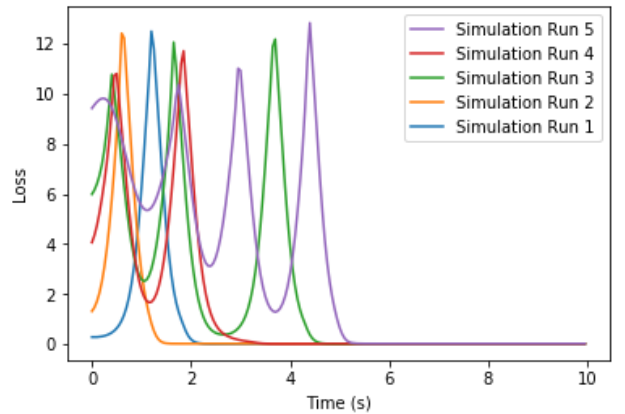

Fig. 9. Effect of random initialization on policy performance.

The total summed loss of rollout is plotted for 100 different starting states sampled randomly from the state-space to show the generalizability of the learned controller in Fig. 10. Less than 15% of simulation run failed to obtain convergence, which can be attributed to extremely high angular velocity retaining its effectiveness.

The real-world bias is introduced by increasing the limits on torque and speed used internally in the simulation with random Gaussian noise scaled to the quantity being modified. The initial state of the environment is kept constant across for equivalent comparison. The results in Fig. 11 shows that the policy learned through simulation works in the real world and achieves convergence in 4 out of 5 cases.

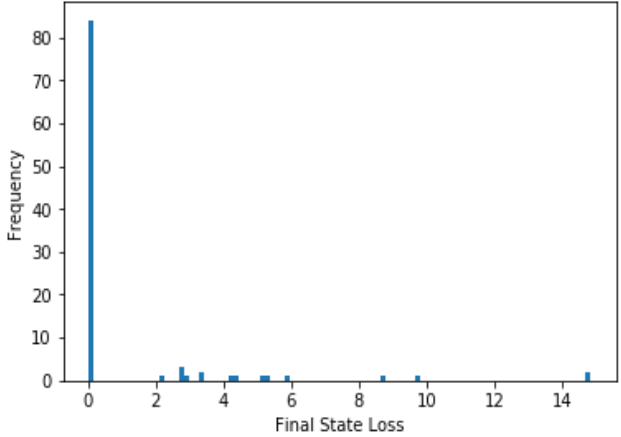

Fig. 10. Generalizability of the learned controller for pendulum environment.

The ablation study for determining the effectiveness of adaptive sampling was conducted and the results are presented in Fig 12 and Fig 13. where AUR-P algorithm was used with and without adaptive sampling. The reward distribution with adaptive sampling has much less variance than without adaptive sampling, leading to the conclusion that adaptive sampling indeed is able to capture and reduce the uncertainty of the dynamic system. For both cases equivalent number of points (125) were used. In the case of adaptive sampling the initial points amounted to 40 and the rest were added as adaptive samples, whereas for without adaptive sampling 125 points were added using LHS.

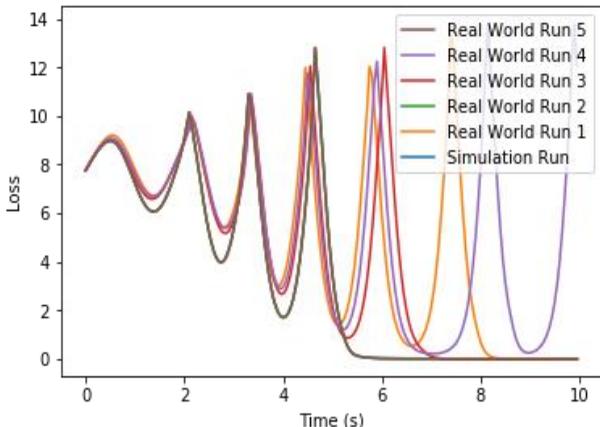

Fig. 11. Generalizability of learned controller for pendulum environment.

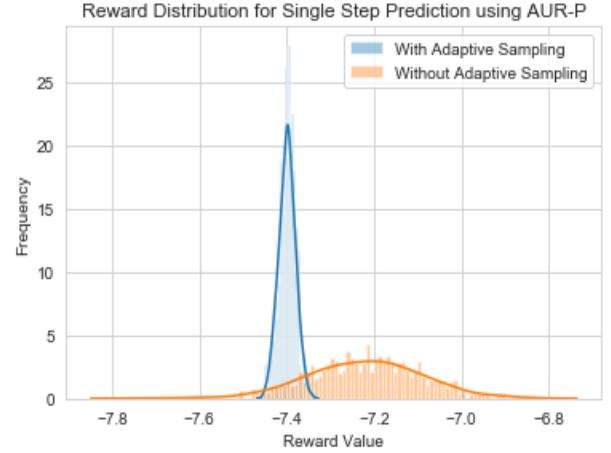

Fig. 12. Ablation study on adaptive sampling: Reward distribution.

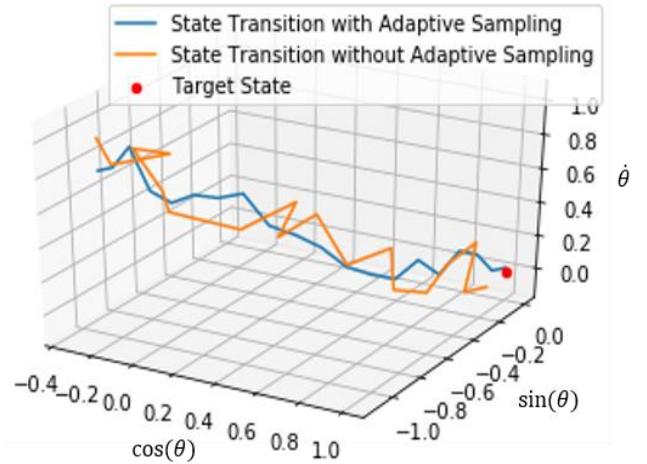

Fig. 13. Ablation study on adaptive sampling: State transition.

It is further shown in Fig. 13 that the controller learned without adaptive uncertainty reduction fails to achieve convergence in the case where a single definite target state exists due to model uncertainty, whereas the controller with adaptive sampling was able to achieve target state with much less path variation.

### 5.2 Cart Pole Balance

The CartPole environment from OpenAI gym consists a pendulum attached to a cart through an un-actuated joint. The cart moves on a frictionless track. At the start, the pole is kept vertical with slight variations in the state parameters drawn from uniform distribution $U(-0.05, 0.05)$. The aim is to keep the cart-pole balanced between the state constraints for more than a preset time. Each trial in the environment is run for at most 200 time steps. The environment is solved when the average time across 100 such trials where the cart-pole remains balanced is at least 195 steps. 100 such trials together are termed as a single run. The mass of the pole is 0.1 kg whereas the mass of the cart is 1 kg. The length of the pole is 1 m.

The observation space consists of 4 continuous states, cart position $x$, cart velocity $\dot{x}$, pole angle $\theta$ and the pole velocity at the tip $\dot{\theta}$. Both the cart velocity and the pole velocity are unbounded. The cart position is bounded within [-4.8m,4.8m] and the pole angle is bounded within [-24°,24°]. The action space consists of a single variable, the direction of force application. The magnitude of the force is constant at 10 N. Direction is limited to two discrete values, towards the right (1) and towards the left (0).

All the steps (till maximum of 200) are marked either as success or failure, all subsequent steps after the first failure are marked as failures. A step is marked as a success if the cart position ≤ ± 2.4m and the pole angle is ≤ ± 12° and failure otherwise. The loss function at any step t is defined as:

$$L(\tilde{x}_t) = 1 - s(\tilde{x}_t) \qquad (16)$$

$$s(\tilde{x}_t) = \begin{cases} 0, & if\ s(\tilde{x}_{t-1}) = 0° \\ 0, if\ x_t > \ \pm 2.4\ or\ \theta_t > \ \pm 12° \\ 1, & otherwise \end{cases} \qquad (17)$$

Internally Euler method is used as solver for Ordinary Differential Equation (ODE) based dynamics. The time difference between two state transitions is 0.02 seconds. Therefore, one trial lasts for a maximum of 4 seconds. The low fidelity virtual environment was created with 50 initial samples and had reward CI value of 0.72.

70 adaptive samples were needed to achieve the target CI value leading to a total of 120 evaluations in the original environment. Since the starting state variation is low, the controller is able to always able to balance the cart-pole in the steady position in 100 trials ran for 100 times. We vary the start state variance from $U(-0.05,0.05)$ to $U(-0.1,0.1)$ to test for in-simulation input state robustness and present the results for 100 runs in Fig 14. With increased input variations, the controller is still able to perform well in 83% of runs where the average score was 200.

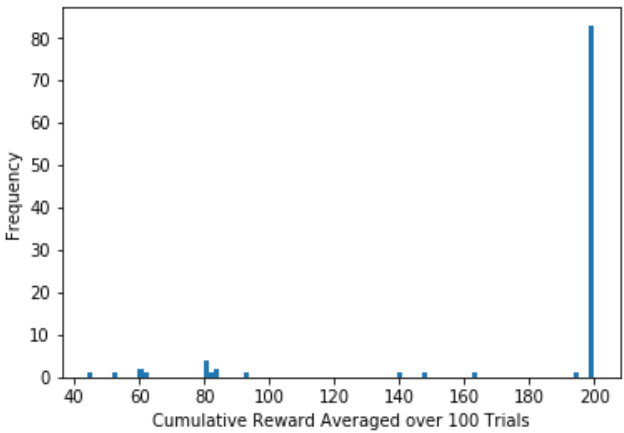

Fig. 14. Generalizability of the learned controller for cartpole environment

We check for performance in the real world by introducing physical uncertainties in the system, the two parameters varied here are the mass of the pole and mass of the cart, changing which significantly changes the dynamics governing the system, thereby imitating bias. This change was again, drawn from a standard normal distribution and scaled to the original dimensions. We found the policy completely impervious to the small changes in the physical dimensions. The effect of physical uncertainty due to the bias in model is presented Fig. 15.

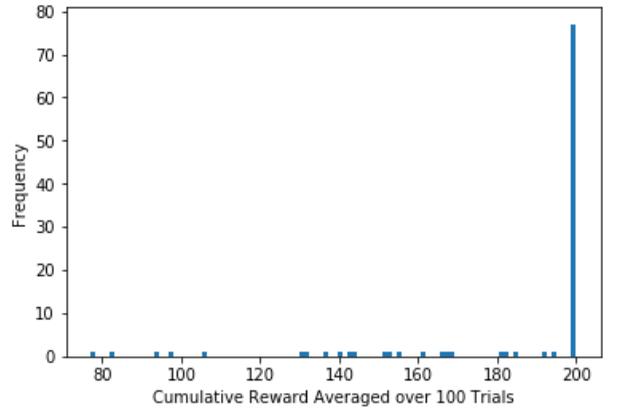

Fig. 15. Effect of physical bias on policy performance for cartpole environment.

## 6 CONCLUSION

AUR gives a practical solution to autonomously find controllers with limited data from real environments (either simulation or real world) and eliminates several drawbacks from model-based methods such as slow offline/planning phase as in [1], [17] and restrictive optimization methods. Our method shows potential to work with high fidelity simulators involving finite element analysis or terrain deformation which is usually avoided when using model free methods due to high number of interactions involved.

While many methods for modelling the dynamics of systems utilize GP at their core, they ignore or limit their usage of uncertainty information inherent to the framework. Our method effectively uses this information for self-improvement and as well as for reward balancing (in AUR-P).

By exploiting advances in GPU architectures, our method is able to scale up the complexity of policy structures involved while keeping the computation time much lower than analytical approaches as well as the previously established model-based methods. As our method focuses on environment as opposed to the agent, it works with any arbitrary policy structure as opposed to [1], which requires computation of gradients analytically.

Due to confidence of model being restricted to only previously visited area, most model based methods have and exploit a small initial state distribution, leading to low robustness. Model free methods on the other hand have increased robustness due to wider reset distribution and high number of random restarts. Our method due to giving importance to global confidence level combines the advantages of both model-based and model free methods, leading to robust controllers with increased data efficiency. The data efficiency in our method can be further attributed to use of Latin hypercube sampling to generate initial samples as opposed to random rollout dependent exploration and sequential sampling to improve CI, leading to only carrying out experimentation at points of maximum expected improvement.

Since the modelling aspect in our method is dependent upon use of Gaussian Processes, our method is restricted by combined state and action dimensionality. However, most problems in robotics and control tend to utilize simplified state space, which makes the use of our method viable. Possible alternatives to this problem include use of Bayesian Neural Networks (BNNs) as described in [10] or modifications on feedforward neural networks to introduce uncertainty information as in[29], [30], or [31].